\documentclass[10pt,twocolumn,letterpaper]{article}

\usepackage{wacv}
\usepackage{times}
\usepackage{epsfig}
\usepackage{graphicx}
\usepackage{amsmath}
\usepackage{amssymb}
\usepackage{graphicx}
\usepackage{dblfloatfix}
\graphicspath{ {./images/} }
\usepackage{capt-of,etoolbox}
\usepackage{etoolbox,lipsum}
\usepackage{blindtext}
\usepackage{multicol}
\usepackage{float}

\usepackage[colorlinks=true, allcolors=blue]{hyperref}
\usepackage{footmisc}
\usepackage{array,multirow}

\usepackage{color,soul}
\usepackage{capt-of,etoolbox}
\pdfoutput=1

\wacvfinalcopy % *** Uncomment this line for the final submission

\def\wacvPaperID{14} % *** Enter the wacv Paper ID here

\ifwacvfinal\fi
\title{A Weakly Supervised Adaptive DenseNet for Classifying Thoracic Diseases and Identifying Abnormalities}

\author{
Bo Zhou \\
School of Computer Science \\
Carnegie Mellon University \\
Pittsburgh, PA, USA \\
\and
Yuemeng Li \\
Department of Bioengineering \\
University of Pennsylvania \\
Philadelphia, PA, USA \\
\and
Jiangcong Wang \\
Department of Bioengineering \\
University of Pennsylvania \\
Philadelphia, PA, USA \\
}

\begin{document}
\twocolumn[{%
\renewcommand\twocolumn[1][]{#1}%
\maketitle

% \begin{center}
%     \centering
%     \includegraphics[width=\textwidth]{bbox_top}
%     \captionof{figure}{Examples of the disease localizations generated from our network. The localization output from our network (red bounding box), that trained only with image-level annotation, match with the ground truth localization (green bounding box).}
%     \label{topfig}
% \end{center}

}]

%\maketitle
\ifwacvfinal\thispagestyle{empty}\fi

%%%%%%%%% ABSTRACT
\begin{abstract}
We present a weakly supervised deep learning model for classifying thoracic diseases and identifying abnormalities in chest radiography. In this work, instead of learning from medical imaging data with region-level annotations, our model was merely trained on imaging data with image-level labels to classify diseases, and is able to identify abnormal image regions simultaneously. Our model consists of a customized pooling structure and an adaptive DenseNet front-end, which can effectively recognize possible disease features for classification and localization tasks. Our method has been validated on the publicly available ChestX-ray14 dataset. Experimental results have demonstrated that our classification and localization prediction performance achieved significant improvement over the previous models on the ChestX-ray14 dataset. In summary, our network can produce accurate disease classification and localization, which can potentially support clinical decisions. 
% \blfootnote{\textsuperscript{\textbf{\dag}} \textbf{Both authors equally contributed to this work}}
\end{abstract}

\section{Introduction}
% general background for CV and CV in medical imaging
Large scale annotated visual datasets have boosted performance of deep learning methods on many challenging computer vision problems \cite{krizhevsky2012imagenet,russakovsky2015imagenet,lin2014microsoft}. Tasks like object detection, classification, tracking, and segmentation have been successfully tackled by techniques built on top of these large-scale dataset with annotations \cite{he2017mask,girshick2015fast,he2016deep}. There are increasing numbers of applications utilizing deep learning methods in medical imaging analysis over the last decade \cite{litjens2017survey}. In clinical procedures, visual evidence such as segmentation or spatial localization of abnormal regions that supports the diagnosis results, is an vital part of clinical diagnosis. This provides a comprehensive interpretation of diagnosis results and potentially decreases the false positive rate.

In this work, we focus on the automatic disease diagnosis and localization in chest radiography released by \cite{wang2017chestx} named ChestX-ray 14, which is one of the largest public chest radiography dataset with image-level disease labels and contains a small subset of region-level disease localizations (bounding boxes). Our goal is to develop a deep learning scheme capable of both classifying the disease and localizing the associated lesion sites. Divergent from standard strongly supervised object detection, our model does not require ground truth localization annotations during training. Firstly, we adopted and modified a pre-trained classification CNN as our front-end feature extractor \cite{huang2017densely}. The pre-trained front-end encodes the information from a large perceptive field. Then, after passing through a simple bridging structure, the extracted feature from the front-end are fed into a customized two-stage pooling network structure, which produces both classification and associated localization simultaneously.

\begin{figure*}[ht]
\centering
\includegraphics[width=\textwidth]{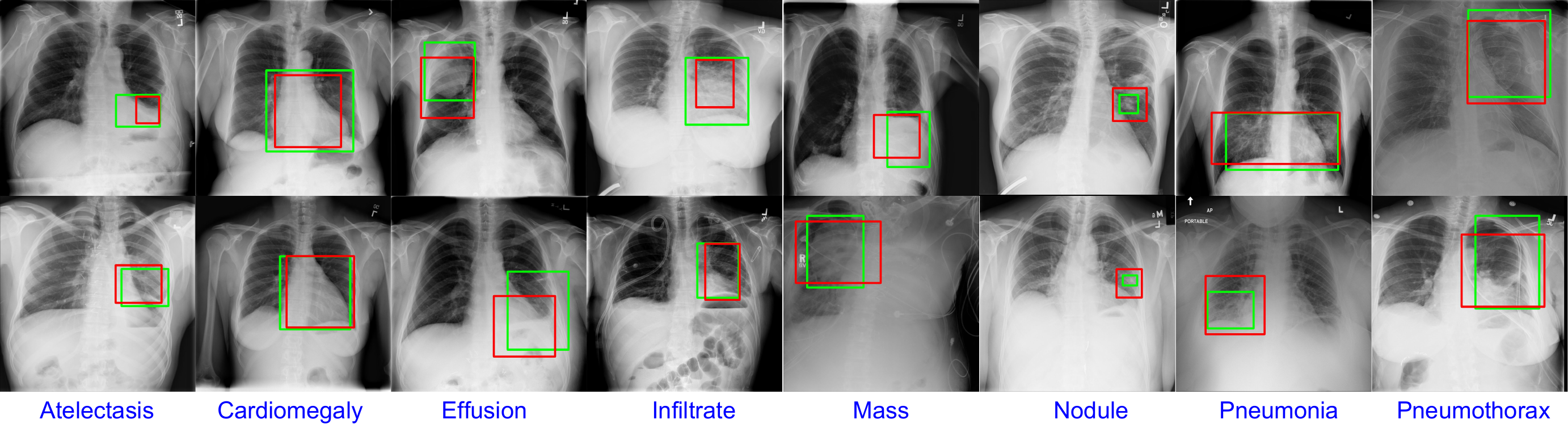}
\caption{Examples of the disease localizations generated from our network. The localization output from our network (red bounding box), that trained only with image-level annotation, match with the ground truth localization (green bounding box).}
\label{topfig}
\end{figure*}

Both quantitative and qualitative visual evaluations show that our proposed model obtains significant improvement over the previous published state-of-the-art results on disease classification and localization. Visual evaluations indicate a strong alignment and correspondence between the clinical annotations and the predicted disease candidate regions shown in figure \ref{topfig}. 

% put a figure to show best results

\noindent \textbf{Contributions:} The contributions of this work are two-folds. First, we developed a weakly supervised end-to-end learning structure that learns from chest radiography images containing multiple common thoracic diseases by explicitly searching over possible disease features and locations in the image. Second, we performed extensive experimental analysis of our model on the large-scale ChestX-ray14 datasets. With the optimal parameter settings found from our analysis, we observed that our model (i) can predict accurate thoracic disease classifications (ii) output better approximate disease locations than previous methods with only image-level labels available. 

\section{Related Work}
\textbf{CAD for chest radiography:} The chest radiography is the most ordered and common radiological examination for chest diseases. It is a low-cost, low radiation and fast imaging exam. Recent studies also demonstrated the chest radiography's application on detection and evaluation of coronary artery diseases, which are usually evaluated using Computed Tomography (CT) with expensive cost and high radiation dose \cite{zhou2017detection,zhou2018visualization,wen2018enhanced}. CAD techniques have been widely applied in chest radiography for task such as automatic diagnosis and patient image retrieval. Previously, Bar \etal adapted the Decaf-Net for 8 thoracic diseases classification on a relatively small chest x-ray dataset \cite{bar2015chest}. Lajhani \etal proposed the DCNN for tuberculosis classification task. Their model ensembled both Alex-Net and Google-Net and achieved impressive results \cite{lakhani2017deep}. Anavi \etal has worked on image retrieval in medicine, specifically for chest radiography given the pathology \cite{anavi2015comparative}. All the aforementioned research showed promising results. However, they only conducted their experiments on relatively small dataset, ranging from 10 to 500 images.

\textbf{ChestX-ray14:} Recently, National Institute of Health (NIH) released one of the largest public chest radiography dataset, consisting of 108,948 posterior-anterior view images from 32,797 patients with eight major chest diseases \cite{wang2017chestx}. A small subset of this dataset is provided with hand labeled bounding boxes for evaluation. Lately, NIH further expanded this dataset to 112,120 frontal-view images with 6 additional thoracic diseases, named ChestX-ray14. Several deep learning methods have been addressed the application of CNN on this dataset for thoracic disease classification and localization \cite{yao2017learning,li2017thoracic,rajpurkar2017chexnet}. Wang \etal \cite{wang2017chestx} applied a pre-trained Res-Net as the backbone to generate heatmap as localization, and subsequently used a global maxpooling to obtain classification. However, most of the localizations generated by their network mismatched with the ground-truth bounding boxes due to only limited extracted feature is used. Yao \etal \cite{yao2017learning} used DenseNet \cite{huang2017densely} to extract features. To harness the correlation between some of the diseases, they used a LSTM module to repeatedly decode the feature vector from a DenseNet \cite{huang2017densely} front end and produced one disease prediction at each step. They achieved improved results compared to the baseline in \cite{wang2017chestx}. One of the most recent work from Rajpurkar \etal achieved a good multi-label classification results by fine-tuning a pre-trained DenseNet-121 \cite{huang2017densely, rajpurkar2017chexnet}. Their classification results outperformed the previous methods \cite{wang2017chestx,yao2017learning,li2017thoracic}. However, there is no localization component and weakly supervision in their work. Another most recent work from Li \etal  \cite{li2017thoracic} used a pre-trained Res-Net to extract features and divided them into patches. They passed the extracted patches through a fully-convolutional classification CNN to obtain a disease probability map which is supervised by both image-level annotations and limited pixel-level annotations. In the mean time, a classification score was acquired by multiplying all probability values together. They achieved significantly better localization results as compared to the baseline \cite{wang2017chestx} for certain diseases. But this approach requires certain amount of bounding box training data for improving their model performance.

\begin{figure*}[!htb]
\centering
\includegraphics[width=0.88\textwidth]{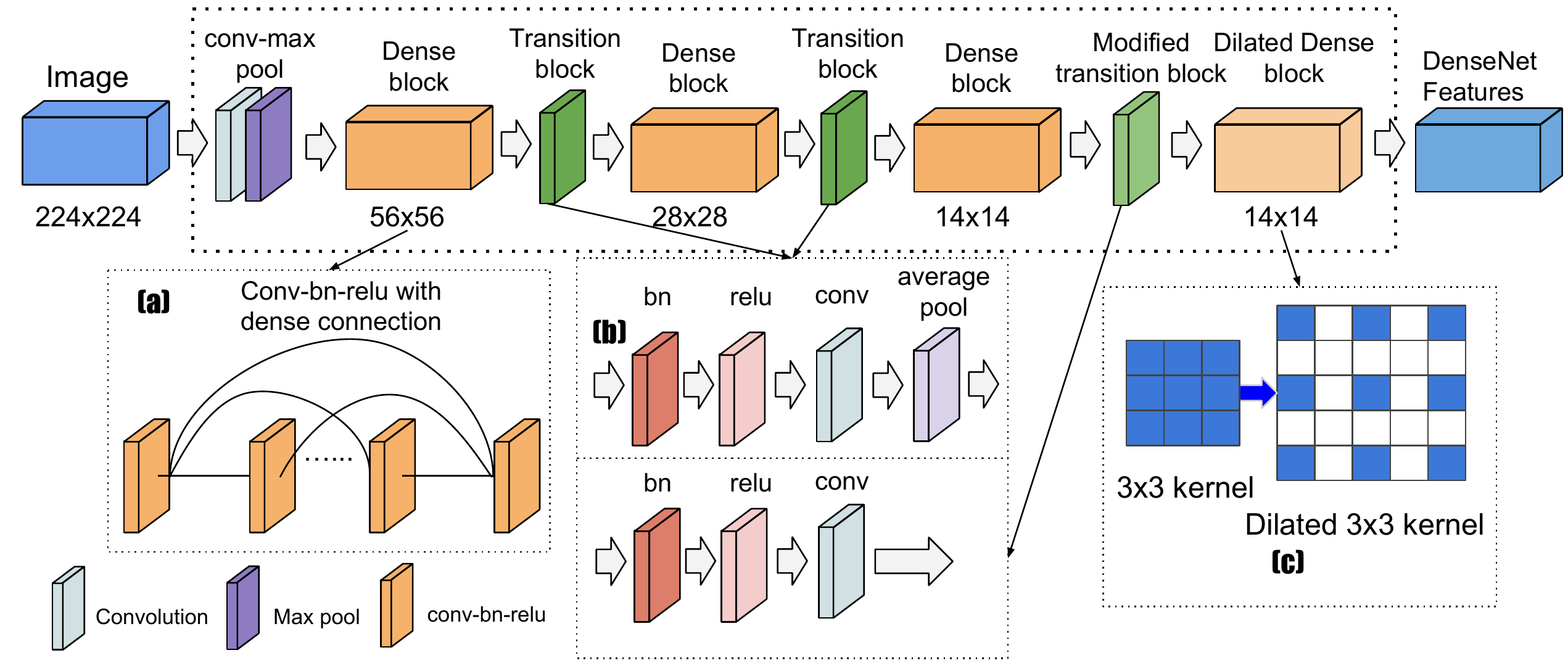}
\caption{The architecture of the adaptive densenet-169 is shown above. The network receives input of size $224 \times 224$, passes it through a convolution and max pool layer, and then $4$ dense blocks interleaved with $3$ transition blocks, and produces $14 \times 14$ feature maps.  
(a) illustrates the dense connection in dense block, where output from each conv-bn-relu layer is concatenated to the input before fed to the next layer. (b) shows the average pool layer is taken out in the modified transition block, so that it maintains same spatial resolution as previous dense block's. (c) illustrates the kernel dilation in dilated dense block. Pre-trained convolution kernels in the last dense block were dilated to keep the receptive field unchanged, after an average pool layer is removed in (b). 
}
\label{adadense}
\end{figure*}

\textbf{Weakly supervised learning:} The ChestX-ray14 dataset \cite{wang2017chestx} contains mostly classification labels but few bounding boxes annotations. Therefore, we seek a model that is capable of localizing the diseased regions given only image wise labels. This falls into category of weakly supervise learning (WSL), which often refers as the task of capturing object location through a customized deep learning model that trained with only image-level labels. Oquab \etal \cite{oquab2015object} proposed a WSL scheme. They used a pre-trained CNN to generate class probability map across spatial locations and applied max-pooling across spatial locations to get a single binary score vector. With their strategy, they were able to get both decent classification accuracy and localization mAP on the benchmark dataset \cite{everingham2011pascal, lin2014microsoft}. Recently, Durand \etal proposed a more sophisticated WSL approach \cite{durand2017wildcat}. They presented the idea of class pooling based on the assumption that an object inherently can be decomposed into different sub-maps to capture better semantic segmentation in WSL. Their experimental results tested on PASCAL 2007, PASCAL 2012, MS COCO, 15 Scene, MIT67 \cite{everingham2008pascal, everingham2011pascal, lin2014microsoft,  quattoni2009recognizing, lazebnik2006beyond} indicated the advantage of such design and achieve significant improvement over \cite{oquab2015object} 's model.

\section{Method}
In this work, we aim to use chest radiographs as input with only image-wise labels to train a model that generates classification for thoracic disease along with the localization heatmaps. The architecture of our proposed network can be summarized into three parts: 1) using an Adaptive DenseNet to generate feature maps and 2) using a bridging layer to convert the feature maps into class-specific sub-maps 3) using a WSL pooling structure to pool the sub-maps into class-specific heatmaps and single probability score for each disease.

%that to pool the sub-maps into class-specific heatmaps and single probability %score for each disease. 

\subsection{Adaptive DenseNet}
The Adaptive DenseNet inherits the basic structure from DenseNet \cite{huang2017densely}. The dense block of DenseNet consists of direct connections from preceding layers to all subsequent layers which pass image information in deep model's training. We maintained the DenseNet structures before the fully connected classification layers as our basic structure. However, the output feature map size after the final dense block is greatly decreased ($7\times7$), losing too much spatial resolution as compared to the input image ($224\times224$), which is not suitable for our localization task. To overcome this, firstly, we took out the average pooling operation at the third transition layer to achieve a fine resolution. Secondly, in order to maintain the same receptive field of each convolution kernel, we dilated all kernels in the fourth dense block. The Adaptive DenseNet outputs a feature map with spatial size of $14\times14$. Details of our Adaptive Densenet is shown in Figure \ref{adadense}. In this work, we chose the DenseNet-169 as our basic structure. Our adaptation can be applied to other DenseNet structures as well, such as DenseNet-121 and Densenet-201. With the Adaptive DenseNet, we were able to acquire a fine resolution feature map.

\vspace*{-0.16cm}
\begin{figure}[!h]
\centering
\includegraphics[width=0.46\textwidth]{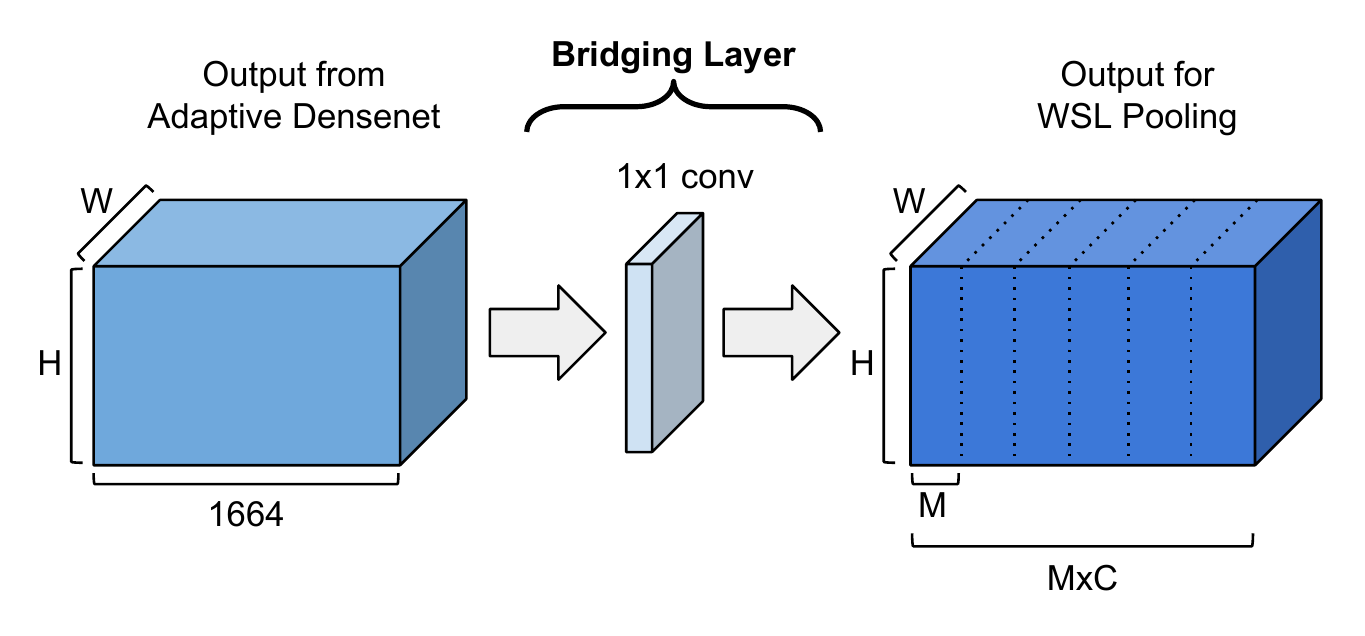}
\caption{Bridging layer connects the output of Adaptive DenseNet and WSL pooling. Given the Adaptive DenseNet output feature, the bridging layer performs a 1x1 convolution to transform the output feature into sub-maps. The output channels of sub-maps are $M\times C$, where $M$ being number of sub-maps per class and $C$ being number of class. The sub-maps fed into the WSL pooling layers.}
\label{bridging}
\end{figure}

\begin{figure*}[!htb]
\centering
\includegraphics[width=0.93\textwidth]{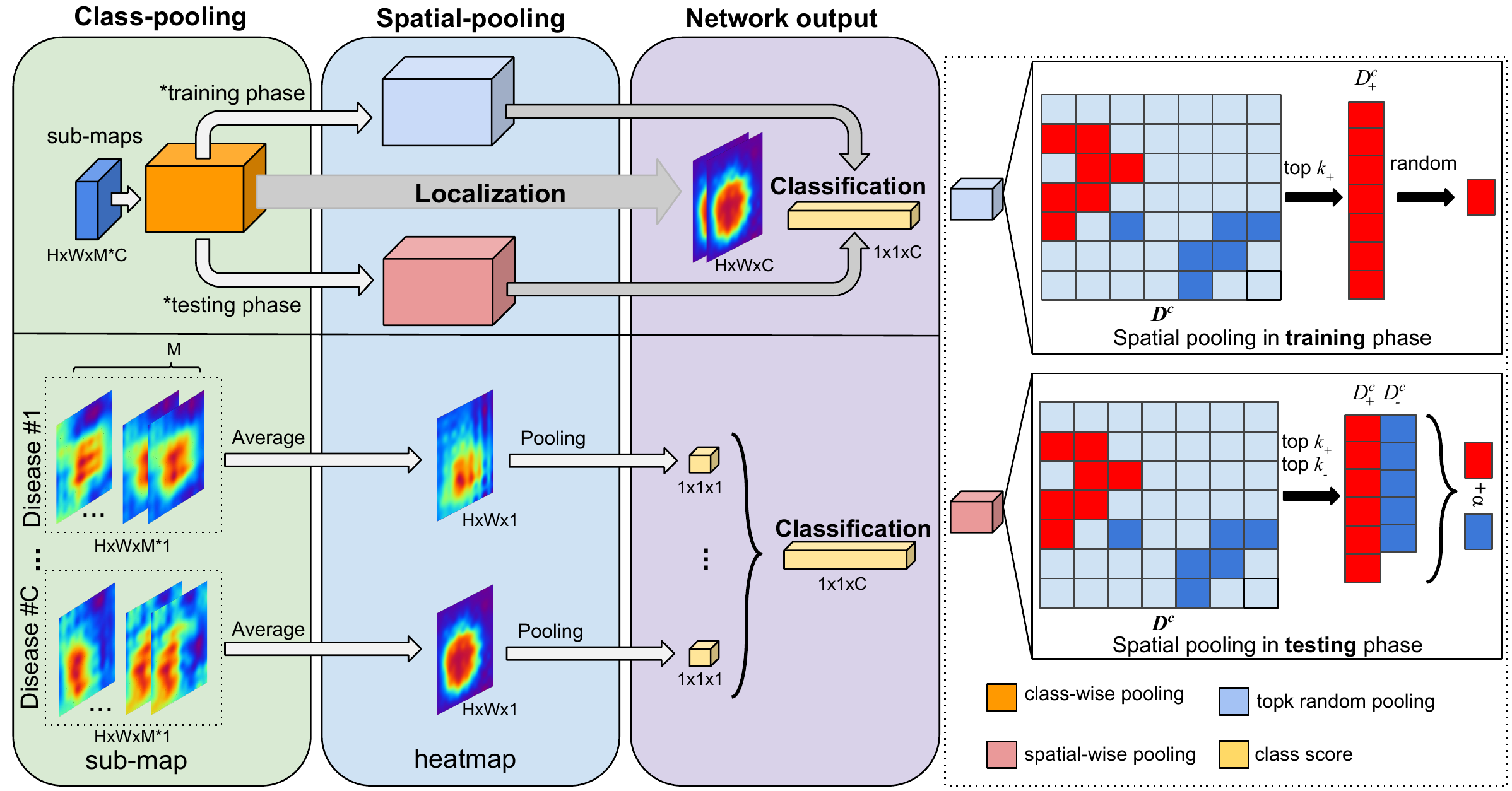}
\caption{Illustration of our 2-stage pooling process: class-wise pooling and spatial pooling. During class-wise pooling (green region), multiple class-specific sub-maps are averaged and a single heatmap is generated for each class. Each classes' heatmap are used for their localization. The heatmap is then passed to spatial pooling. We employed two different spatial pooling schemes during training and testing. During training, single random score from top $k_+$ scores in the heatmap is selected to increase robustness against erroneous top scores. During testing, top $k_+$ scores and bottom $k_-$ scores are weighted averaged with weight on bottom samples to give maximum performance. The final output of spatial pooling is a single class score for each class.}
\label{pooling}
\end{figure*}

\subsection{Bridging layer}
To connect the output of the Adaptive DenseNet and the pooling structure described in section \ref{section_wslpooling}, a bridging layer is proposed and used here. Specifically, the output of the Adaptive DenseNet is transformed by a $1\times 1$ convolutional layer, which learns the mapping from the Adaptive DenseNet's output feature maps to class specific sub-maps. Through the bridging layer, the channel size of the feature map is transferred from $1664$ (channel size of Adaptive DenseNet's output) to $M \times C$, as shown in Figure \ref{bridging}. This bridging layer connects the Adaptive DenseNet and the WSL pooling structure together to form the overall network. 

\subsection{WSL pooling}
\label{section_wslpooling}
To obtain accurate disease classification along with disease localization heatmap, we integrated a customized weakly supervised learning strategy into our network. Similar to \cite{durand2017wildcat}, we implemented two stages of global pooling layers to summarize all information for each class in feature map outputted from the Adaptive DenseNet, which include: 1) a class-wise pooling within the same class, and 2) a spatial pooling along different classes. The class-wise pooling summarizes submaps into final localization heatmap, and the spatial-wise pooling summarizes heatmap into classification. Details of the pooling designs are demonstrated in Figure \ref{pooling}.

\textbf{class-wise pooling:} The class-wise pooling combines the $M$ submaps from each of the $C$ classes from the bridging layer. In specific, our class-wise pooling combines the $M$ maps for all disease classes independently through Equation \ref{classwise}. 
\begin{equation}
\bar{D}_{i,j}^c = \frac{1}{M}\sum\limits_{m=1}^M D_{i,j}^{c,m}
\label{classwise}
\end{equation}
where $M$ is the number of sub-maps with the assumption that each sub-map contains different features of the corresponding disease. Unlike traditional WSL with only one map considering only one extracted feature, our multiple maps strategy is useful for the disease classification/localization tasks by utilizing multiple extracted features. $H$, $W$, $C$ are feature map's height, width, and the number of class, respectively. The input feature $D$ with size of $H\times W\times MC$ is divided into $C$ branches with each size of $H\times W\times M$. Then, the sub-maps information are composed into one final feature map for each class through an average pooling. This results in a transformed class heatmap of size $H\times W\times C$.

\textbf{spatial-wise pooling:} 
Given a class heatmap with size of $H\times W\times C$, we used a spatial-wise pooling to generate our final classification output. Our spatial-wise pooling extracts a classification vector ($1\times 1\times C$) from the class heatmap  outputted from class-wise pooling shown above. The most common strategy for spatial-wise pooling is maxpooling, which might potentially ignore most of the feature information, generate localization heatmap only in small regions, and cause insufficient pass of gradient for training. 

Therefore, we proposed to use a customized spatial-wise pooling strategy for both training and testing of our network, as shown in Figure \ref{pooling}. During the training phase, we firstly selected top $k$ scores from $H\times W$ heatmaps and randomly chose one among the k numbers for each class. Because the class heatmap may not be perfect, the highest score may not correctly correspond to the disease location. Thus, random sampling from top $k$ spatial locations instead of simply choosing the max gives the network higher probability to capture the correct spatial location of the disease and generate the correct gradient. During the testing phase, we looked at regions with the highest($D_+$)/lowest($D_-$) activation from the heatmap $D$. The highest activation $D_+$ indicates presence of classes while the lowest activation $D_-$ indicates absence of classes. Thus both are incorporated to the prediction layer to achieve a robust classification performance. We introduced a weighting factor $\alpha$ to control the relative importance between these two terms.

Formally, let $D^c$ be the heatmap for class $c$, let $D^c_+$/ $D^c_-$ be the set of top $k_+$/bottom $k_-$ scores in $D^c$, and let $d^c_{final}$ be the final class score. During training phase, we have:
\begin{equation}
d^c_{final} = dr_+, 
\end{equation}
where $dr_+\in D^c_+$ is a single element uniformly random sampled from $D^c_+$.  During testing phase, we have:
\begin{equation}
d^c_{final} = \frac{1}{k_+} \sum\limits_{d_+\in D^c_+} d_+ + \alpha(\frac{1}{k_-}\sum\limits_{d_-\in D^c_-} d_-)
\end{equation}
where $d_+$/$d_-$ are elements in $D^c_+$/$D^c_-$.  and $\alpha$ is the weighting factor added on bottom scores.

\subsection{Network's training}
For network initialization, we transfered the weights from the pre-trained models on ImageNet for our Adaptive DenseNet. Then, we randomly initialized parameters for bridging and WSL pooling layers. We used learning rate of 0.002 with weight decay of 0.1 for every 10 epochs during training. Our model were implemented with Pytorch (\href{https://pytorch.org/}{https://pytorch.org/}) on a Nvidia GTX 1080Ti. We added batch normalization \cite{ioffe2015batch} after each convolution and additional drop out of rate of 0.1 for each DenseBlock inside Adaptive DenseNet \cite{tinto1975dropout} to prevent overfitting. Since there is a large data imbalance between each class, we trained our model with weighted binary cross entropy loss: 
\begin{equation}
\begin{aligned}
L(X,y) = -w_+ \cdot ylogp(Y=1|X) - \\ w_- \cdot (1-y)logp(Y=0|X)
\end{aligned}
\end{equation}
where $w_+$ represents percent of positive sample and $w_-$ represents percent of negative sample among all dataset. 

\subsection{Disease classification and localization}
The heatmap obtained from our class-wise pooling is further used as a reference map to generate disease bounding boxes. A thresholding followed by connected component analysis are applied for each disease class to generate bounding boxes. Empirically, we used a threshold value of 0.8 for "Cardiomegaly" and threshold values of 0.9 for the rest of classes to obtain the best results. The disease classification score is acquired from the spatial-wise pooling on the heatmap. In short, our network structure is designed for both disease classification and localization tasks, yet only use chest radiographs along with image-level annotations for our training. 

\section{Experiments and Results}
\subsection{Data for experiments}
We performed our experiment using the ChestX-ray14 dataset. The diseases in the corresponding chest radiograph are diagnosed by radiologists and the labeled ground truth are obtained through text mining on the patients' diagnostic reports. We used exactly same published data split as in \cite{rajpurkar2017chexnet,wang2017chestx,yao2017learning}. The dataset is divided into training (70\%), validation (10\%) and testing (20\%). The original resolution of image is $1024 \times 1024$ and was downscaled to $256 \times 256$. In order to fit dataset into ImageNet \cite{krizhevsky2012imagenet} pre-trained models, we normalized the image by mean and standard deviation of the images from ImageNet. We used random crop of size  $224 \times 224$ from the downscaled image as the network input for training and we used center crop of same size for testing. For localization validation, the ChestX-ray14 dataset contains 880 images with 983 disease bounding boxes annotated by board-certified radiologists. We calculated the intersection over union (IoU) between our prediction and the ground truth bounding boxes.

\subsection{Disease classification}

%table for result section
\begin{table*}[!htp]
  \centering
  \begin{tabular}{| m{3cm} | >{\centering\arraybackslash}m{2.3cm} | >{\centering\arraybackslash}m{2.3cm} | >{\centering\arraybackslash}m{2.1cm} | >{\centering\arraybackslash}m{2.8cm} | >{\centering\arraybackslash}m{2.3cm} |}
   \hline
   Pathology		& Wang \cite{wang2017chestx}	& Yao \cite{yao2017learning} & Li \cite{li2017thoracic} & Rajpurkar \cite{rajpurkar2017chexnet} & Ours \\
   \hline\hline
   Atelectasis 		& 0.7003	& 0.772 & 0.80  & 0.8094	& \textbf{0.8121} \\
   Cardiomegaly 	& 0.8100  	& 0.904 & 0.87 	& \textbf{0.9248}	& 0.9066 \\
   Effusion 		& 0.7585 	& 0.859	& 0.87 	& 0.8638	& \textbf{0.8786} \\
   Infiltration 	& 0.6614 	& 0.695	& 0.70 	& \textbf{0.7345}	& 0.7065 \\
   Mass 			& 0.6933 	& 0.792	& 0.83 	& \textbf{0.8676}	& 0.8354 \\
   Nodule 			& 0.6687  	& 0.717	& 0.75 	& 0.7802	& \textbf{0.7852} \\
   Pneumonia 		& 0.6580 	& 0.713	& 0.67 	& 0.7680	& \textbf{0.7810} \\
   Pneumothorax 	& 0.7993	& 0.841	& 0.87 	& 0.8887	& \textbf{0.8911} \\
   Consolidation 	& 0.7032	& 0.788	& 0.80 	& 0.7901	& \textbf{0.8115} \\
   Edema 			& 0.8052	& 0.882	& 0.88 	& 0.8878	& \textbf{0.8953} \\
   Emphysema		& 0.8330		& 0.829	& 0.91 	& 0.9371	& \textbf{0.9373} \\
   Fibrosis			& 0.7859	& 0.767	& 0.78 	& 0.8047	& \textbf{0.8187} \\
   Pleural Thickening & 0.6835	& 0.765	& \textbf{0.79} 	& \textbf{0.8062}	& 0.7792 \\
   Hernia 			& 0.8717	& 0.914	& 0.77 	& 0.9164	& \textbf{0.9487} \\
   \hline
  \end{tabular}
  \caption{AUROC is used to evaluate model classification performance for 14 diseases. The best results are shown in bold text. In order to obtain a comprehensive comparison, performance of all previous networks to our knowledge are included. All results are obtained from their latest update.}
  \label{result1}
\end{table*}

We calculated the Area under Receiver Operating Characteristic curve (AUROC) for each class to evaluate the classification performance of our model. Our results are selected based on the best performance of classifier for each class throughout our training iterations. Our classification performance is compared with other state-of-the-art results \cite{wang2017chestx,yao2017learning,li2017thoracic,rajpurkar2017chexnet}. The experiment is tested with the same dataset splitting as demonstrated in \cite{wang2017chestx,yao2017learning,rajpurkar2017chexnet}.

\begin{figure*}[!htb]
\centering
\includegraphics[width=\textwidth]{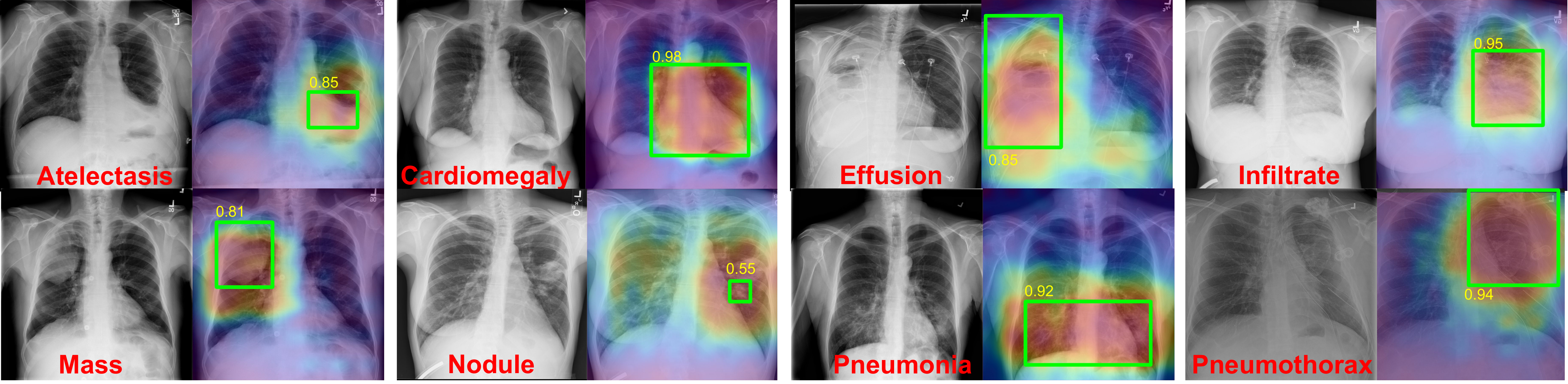}
\caption{Examples of localization results on 8 diseases along with ground truth bounding box annotations. In each pair, a chest radiograph (left) inputs into our model and the corresponding heatmap (right) is generated. The heatmaps produced from our model match with the ground-truth bounding boxes (green) annotated by radiologists and indicated good IoU results (yellow).}
\label{bbox}
\end{figure*}

In all 14 diseases, our model achieved better classification AUROC than \cite{wang2017chestx,yao2017learning} as shown in Table \ref{result1}. In comparison to \cite{li2017thoracic} with supervision from both image-level labels and bounding box annotations, our model achieved significant better classification AUROC in 13 out of 14 diseases. In particular, our model obtained more than 3\% higher AUROC in "Cardiomegaly", "Nodule", "Pneumonia", "Fibrosis" and "Hernia". Comparing our results with \cite{rajpurkar2017chexnet}'s work which only focus on classification, our model outperformed theirs in 10 out of 14 diseases.

In Table \ref{result1}, our network performed better on diseases classification with large lesions than one with smaller lesions. "Cardiomegaly" and "Emphysema" are the representative classes for disease with large lesion and our model achieved classification score over 90\% on both. Although our network performed less accurate on identifying smaller lesions such as mass and nodule, our model is improved from \cite{wang2017chestx,yao2017learning,li2017thoracic,rajpurkar2017chexnet} on these classes as well. Unlike the traditional maxpooling which only output the maximal response from a network output, our class-wise pooling can pool multiple response from the submaps which contain different feature responses of the target lesion. In this case, the classification of the large lesions that potentially contains more appearance feature than small lesions can be better improved with our customized pooling methods.

In addition, we performed hyper-parameter searching for best classification performance (details shown in Supplemental Materials). In class-wise pooling, we tested number of sub-maps ($M$) from number 2 to 18 and found 14 sub-maps yielded best result. We then kept this setting for the rest of experiment. In spatial pooling, we tested hyper-parameters on two different strategies separately. During training, we tested our model using top $k_+$ scores, $k_+$ ranging from 1 to 20. We found that $k_+$ at 10 yielded best network. During testing, we tested $k_+$ from 1 to 20 and $k_-$ from 1 to 25. We found that both $k_+$ and $k_-$ at 15 yielded best classification result. We tested $\alpha$ ranging from 0.25 to 1 and $\alpha$ equals at 1 gave the best importance balance between $k_+$ and $k_-$. 

\subsection{Disease localization}

%% Table for bounding box performance-average iou+T(ious)
\begin{table*}[!htb]
\centering
\begin{tabular}{ >{\centering\arraybackslash}p{1cm} | >{\centering\arraybackslash}p{1cm} | >{\centering\arraybackslash}p{1.4cm} >{\centering\arraybackslash}p{1.8cm} >{\centering\arraybackslash}p{1.1cm} >{\centering\arraybackslash}p{1.3cm} >{\centering\arraybackslash}p{0.9cm} >{\centering\arraybackslash}p{1.1cm} >{\centering\arraybackslash}p{1.4cm} >{\centering\arraybackslash}p{1.6cm} }
\hline
T(IoU)		& Model 		& Atelectasis 		& Cardiomegaly 		& Effusion 		& Infiltration 		& Mass 		& Nodule 		& Pneumonia 		& Pneumothorax 		\\
\hline
% Row1
\multirow{3}{*}{0.1} 
&Wang & \textbf{0.6888} & 0.9383 & \textbf{0.6601} & 0.7073 & 0.4000 & \textbf{0.1392} & 0.6333 & \textbf{0.3775}\\%\cline{2-10}
%&Li & 0.71 & 0.98 & 0.87 & 0.92 & 0.71 & 0.40 & 0.60 & 0.63\\\cline{2-10}
&Ours & 0.4113 & \textbf{0.9659} & 0.5950 & \textbf{0.8134} & \textbf{0.5292} & 0.0758 & \textbf{0.7421} & 0.3267\\%\cline{2-10}
&Ours* & 0.4283 & 0.9657 & 0.6926 & 0.8211 & 0.5539 & 0.2156 & 0.7752 & 0.3263\\\cline{1-10}

% Row2
\multirow{3}{*}{0.2} 
&Wang & \textbf{0.4722} & 0.6849 & \textbf{0.4509} & 0.4796 & 0.2588 & \textbf{0.0506} & 0.35 & \textbf{0.2346}\\%\cline{2-10}
%&Li & 0.53 & 0.97 & 0.76 & 0.83 & 0.59 & 0.29 & 0.50 & 0.51\\\cline{2-10}
&Ours & 0.2388 & \textbf{0.9453} & 0.3341 & \textbf{0.6745} & \textbf{0.3058} & 0.0125 & \textbf{0.5752} & 0.2043\\%\cline{2-10}
&Ours* & 0.2557 & 0.9453 & 0.5036 & 0.6833 & 0.3292 & 0.1522 & 0.6331 & 0.2052\\\cline{1-10}

% Row3
\multirow{3}{*}{0.3} 
&Wang & \textbf{0.2444} & 0.4589 & \textbf{0.3006} & 0.2764 & 0.1529 & \textbf{0.0379} & 0.1666 & 0.1326\\%\cline{2-10}
%&Li & 0.36 & 0.94 & 0.56 & 0.66 & 0.45 & 0.17 & 0.39 & 0.44\\\cline{2-10}
&Ours & 0.1458 & \textbf{0.9316} & 0.1448 & \textbf{0.5223} & 0.1531 & 0.0031 & \textbf{0.4420} & \textbf{0.1432}\\%\cline{2-10}
&Ours* & 0.1613 & 0.9246 & 0.3334 & 0.5368 & 0.1768 & 0.1393 & 0.5010 & 0.1432\\\cline{1-10}

% Row4
\multirow{3}{*}{0.4} 
&Wang & \textbf{0.0944} & 0.2808 & \textbf{0.2026} & 0.1219 & 0.0705 & \textbf{0.0126} & 0.075 & 0.0714\\%\cline{2-10}
%&Li & 0.05 & 0.18 & 0.11 & 0.07 & 0.01 & 0.01 & 0.03 & 0.03\\\cline{2-10}
&Ours & 0.0887 & \textbf{0.8772} & 0.0590 & \textbf{0.3575} & \textbf{0.1174} & 0.0025 & \textbf{0.2582} & \textbf{0.1124}\\%\cline{2-10}
&Ours* & 0.1062 & 0.8769 & 0.2678 & 0.3911 & 0.1430 & 0.1388 & 0.3252 & 0.1123\\\cline{1-10}

% Row5
\multirow{3}{*}{0.5} 
&Wang & \textbf{0.0500} & 0.1780 & \textbf{0.1111} & 0.0650 & 0.0117 & \textbf{0.0126} & 0.0333  & 0.0306\\%\cline{2-10}
%&Li & 0.14 & 0.84 & 0.22 & 0.30 & 0.22 & 0.07 & 0.17 & 0.19\\\cline{2-10}
&Ours & 0.0448 & \textbf{0.7806} & 0.0133 & \textbf{0.2523} & \textbf{0.0592} & 0.00022 & \textbf{0.1419} & \textbf{0.0514}\\%\cline{2-10}
&Ours* & 0.0612 & 0.7812 & 0.2286 & 0.2843 & 0.0827 & 0.1394 & 0.2086 & 0.0522\\\cline{1-10}

% Row6
\multirow{3}{*}{0.6} 
&Wang & \textbf{0.0222} & 0.0753 & \textbf{0.0457} & 0.0243 & 0.0000 & \textbf{0.0126} & 0.0166 & 0.0306\\%\cline{2-10}
%&Li & 0.07 & 0.73 & 0.15 & 0.18 & 0.16 & 0.03 & 0.10 & 0.12\\\cline{2-10}
&Ours & 0.0116 & \textbf{0.5211} & 0.0000 & \textbf{0.1552} & \textbf{0.0238} & 0.0000 & \textbf{0.0422} & \textbf{0.0413}\\%\cline{2-10}
&Ours* & 0.0280 & 0.5208 & 0.2159 & 0.1873 & 0.0475 & 0.1393 & 0.1081 & 0.0409\\\cline{1-10}

% Row7
\multirow{3}{*}{0.7} 
&Wang & 0.0055 & 0.0273 & \textbf{0.0196} & 0.0000 & 0.0000 & 0.0000 & 0.0083 & \textbf{0.0204}\\%\cline{2-10}
%&Li & 0.04 & 0.52 & 0.07 & 0.09 & 0.11 & 0.01 & 0.05 & 0.05\\\cline{2-10}
&Ours & \textbf{0.0063} & \textbf{0.2342} & 0.0000 & \textbf{0.0742} & \textbf{0.0123} & 0.0000 & \textbf{0.0165} & 0.0108\\%\cline{2-10}
&Ours* & 0.0223 & 0.2332 & 0.2161 & 0.1062 & 0.0358 & 0.1395 & 0.0836 & 0.0104\\\cline{1-10}

% Row8
\hline\hline
\multirow{3}{*}{IoU}
&Wang & 0.2254 & 0.3776 & 0.2558 & 0.2392 & 0.1277 & 0.0379 & 0.1833 & 0.1282 \\
&Ours & 0.1353 & 0.7508 & 0.1637 & 0.4071 & 0.1715 & 0.0134 & 0.3169 & 0.1272 \\
&Ours* & 0.1519 & 0.7497 & 0.3511 & 0.4300 & 0.1956 & 0.1520 & 0.3764 & 0.1272\\\cline{1-10}
\end{tabular}
\caption{Comparison of disease localization using IoU, where threshold of IoU are set to 0.1, 0.2, 0.3, 0.4, 0.5, 0.6, 0.7 for evaluation, respectively. Best results between Wang \etal's and ours are shown in \textbf {bold}. Last row shows average IoU for all diseases. * means results obtained from same dataset with expanded annotation from certified radiologist.}
\label{result2}
\end{table*}

We generated our predicted bounding boxes by applying a naive thresholding ($T=0.1\sim0.7$) on the normalized heatmap with value ranged from 0 to 1 that obtained from our class-wise pooling layer. Examples are shown in Figure \ref{bbox}. No localization annotation was used during training process. The heatmaps were generated by our network with training only from image-level labels. All 880 images with bounding boxes ground truth were used for evaluation.  

In Table \ref{result2}, we compared our results with \cite{wang2017chestx}. Our model achieved significantly higher IoU score over \cite{wang2017chestx} on diseases with large lesion, such as "Cardiomegaly" (abnormally large heart). The heatmap is well-fitted with the ground truth bounding box. Disease such as "Infiltration" (substance such as blood infiltrates through vessel into lung) and "Pneumonia" (inflammation in lung) that affect large area of lungs manifest as visually more prominent patterns, which are learned by our network. Moreover, the localization for small disease region such as "Atelectasis" (partially collapsed lung) and Mass (an abnormal lump $>$ 3cm) can also be well captured by our network. Noted that for "Effusion" (liquid occupying lung space), our network detects the disease that covers whole lesion while the ground truth includes some extra parts such as the shoulder. 

\begin{figure*}[!htb]
\centering
\includegraphics[width=0.95\textwidth]{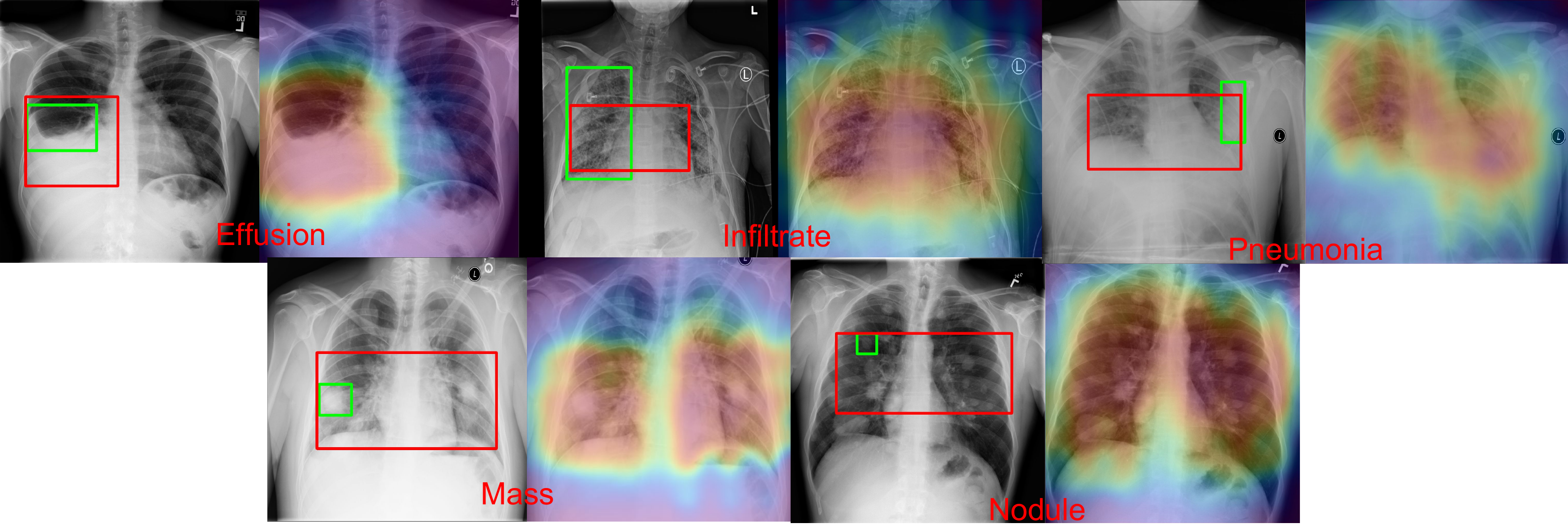}
\caption{This shows 5 ambiguous cases on 5 diseases where bounding annotations are bias. The input image is shown on the right, with two bounding boxes labels: green is the provided ground truth and red is our localized bounding boxes. The class-pooling output heatmap is shown on the right. The red bounding boxes are generated by applying a naive thresholding.}
\label{specialcase}
\end{figure*}

We presents 5 ambiguous cases on 5 diseases in Figure \ref{specialcase}, which represents bias in localization annotations that leads to lower IoU score. In the case of "Effusion", the annotation outlined the liquid-lung boundary. Instead, our network includes the full liquid rinsed area. In the cases of "Infiltration" and "Pneumonia" that spread both lungs, the ground truth annotations only includes single lung, whereas our network captures both lungs. For cases like "Mass" and "Nodule", the ground truth bounding box only highlights one of many instances of "Mass" and "Nodule", but our localization highlights all instances. 

Observing annotation ambiguity, we attempted to quantify the effect. We expanded the annotations by having a certified radiologist manually re-labeled bounding boxes on lesion on 80 ambiguous cases. We evaluated our model with the expanded annotations and the updated localization IoU is shown in Table \ref{result2}. The updated IoU score on expanded annotations is significantly improved for certain classes, such as "Effusion",  "Nodule",  "Atelectasis",  "Infiltration",  "Mass",  and "Pneumonia".

% Image for bounding boxes special case

\section{Discussion}
In clinical procedures, visual evidence such as segmentation or spatial localization of disease lesions, in support of disease classification results, is a vital part of clinical diagnosis. It provides a comprehensive insight into the disease and potentially decreases the false positive of diagnosis. In this work, we proposed a weakly supervised adaptive DenseNet architecture. It only trains on image-level annotation, yet is able to provide both disease classification and corresponding visual evidence, making it potentially valuable in clinical setting. 

In our experiment, we specifically looked at 14 different thoracic diseases. Our network demonstrated its ability to precisely identify disease patterns which generated accurate disease classification and corresponding heatmaps for disease localization. In our classification experiments, our network outperformed the current state-of-the-art method on 10 out of 14 diseases. Specifically, our network has shown significant classification improvements on diseases with large lesion such as "Pneumonia" and "Emphysema". Our model also gave robust classification results on diseases with small lesions such as "Nodule". In our localization experiments, our network achieved significant better localization performance on 5 out of 8 diseases with mean T(IoU)=$0.7$, as compared to the NIH baseline \cite{wang2017chestx}. Compared to previous methods, the possible reasons for our network to achieve higher classification and localization accuracy is two-fold. First of all, the use of a two-stage pooling (class-wise \& spatial-wise) allows the network to capture complex intra-class variations and highlight different useful lesion features on multiple sub-maps for one disease class. Secondly, the removal of average pooling helps to improve spatial resolution and maintain more image details for the lesion features capturing in the WSL pooling.

We also evaluated the effect of localization label ambiguity. We collaborated with a certified radiologist to expand the ground truth bounding box annotations on 80 cases. We found that our localization has significantly better overlap with the bounding box in the expanded annotation than the one provided in ChestX-ray14 dataset on certain cases. Examples of the ambiguity cases are illustrated in our Supplemental Materials with detailed explanations. 

For future work, it is worth noting that the annotations in the ChestX-ray14 dataset is generated by text mining instead of manual annotation, which guarantees 90\% correctness of labeling. In this case, the labels can potentially be further improved. The model should be able to achieve enhanced performance if such training data is available. Furthermore, a dataset with more disease categories included will be favorable as it is closer to real life clinic settings. Future works include expanding the current ChestX-ray14 dataset with more manual label and more disease categories such as coronary artery disease \cite{zhou2017detection,zhou2018visualization} for evaluation. Applications of our model on different medical imaging dataset that targeting multiple diseases should also be evaluated in the future.

\section{Conclusion}
In summary, we presented a weakly supervised adaptive DenseNet to classify and localize 14 thoracic diseases by using only image-level annotation during training. Extensive experiments demonstrated the effectiveness of our network which achieved both the best classification and localization results among the previous published state-of-the-art approaches on the ChestX-ray14 dataset.

\clearpage

{\small
\bibliographystyle{unsrt}
\bibliography{egbib}
}

\end{document}